\pdfoutput=1

\documentclass[11pt,a4paper]{article}
\usepackage{graphicx}

\usepackage[]{emnlp2021}
\usepackage{graphicx}

\usepackage{times}
\usepackage{booktabs}
\usepackage{latexsym}
\usepackage{enumitem}
\usepackage{subfiles}
\usepackage{tabularx}
\usepackage{multirow}
\usepackage[T1]{fontenc}

\usepackage[utf8]{inputenc}

\usepackage{microtype}

\setlist[itemize]{itemsep=1pt, topsep=1pt, leftmargin=13pt}
\setlist[enumerate]{itemsep=1pt, topsep=1pt, leftmargin=13pt}
\title{Cross-Domain Data Integration for Named Entity Disambiguation in Biomedical Text}

\author{Maya Varma \\
 Stanford University\\
 \small\texttt{mvarma2@cs.stanford.edu} \And
 Laurel Orr\\
 Stanford University\\
 \small\texttt{lorr1@cs.stanford.edu} \And 
 Sen Wu\\
 Stanford University\\
 \small\texttt{senwu@cs.stanford.edu} \AND
 Megan Leszczynski\\
 Stanford University\\
 \small\texttt{mleszczy@cs.stanford.edu} \And
 Xiao Ling\\
 Apple\\
 \small\texttt{xiaoling@apple.com} \And 
 Christopher Ré\\
 Stanford University\\
 \small\texttt{chrismre@cs.stanford.edu}}

\begin{document}
\maketitle
\begin{abstract}
Named entity disambiguation (NED), which involves mapping textual mentions to structured entities, is particularly challenging in the medical domain due to the presence of rare entities. Existing approaches are limited by the presence of coarse-grained structural resources in biomedical knowledge bases as well as the use of training datasets that provide low coverage over uncommon resources. In this work, we address these issues by proposing a cross-domain data integration method that transfers structural knowledge from a general text knowledge base to the medical domain. We utilize our integration scheme to augment structural resources and generate a large biomedical NED dataset for pretraining. Our pretrained model with injected structural knowledge achieves state-of-the-art performance on two benchmark medical NED datasets: MedMentions and BC5CDR. Furthermore, we improve disambiguation of rare entities by up to 57 accuracy points.

\end{abstract}

\section{Introduction}
\label{sec:intro}
\begin{figure*}
\centering
\includegraphics[width=\textwidth]{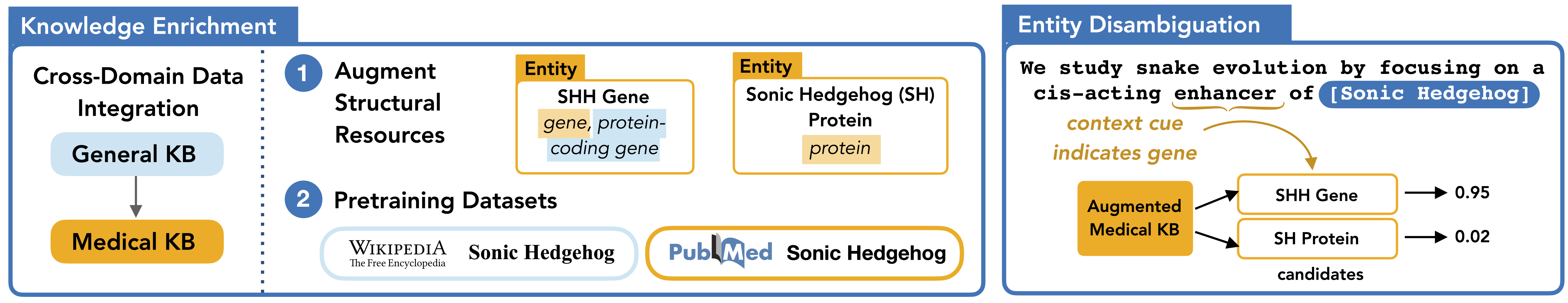}
\caption{\label{citation-guide}
(Left) We integrate structural knowledge between a general text KB and a medical KB, which allows us to augment structural resources for medical entities and generate pretraining datasets. (Right) A pretrained model injected with augmented structural information can now better reason over context cues to perform NED.
}

\end{figure*}
Named entity disambiguation (NED), which involves mapping mentions in unstructured text to a structured knowledge base (KB), is a critical preprocessing step in biomedical text parsing pipelines \cite{percha2020}. For instance, consider the following sentence: “We study snake evolution by focusing on a cis-acting enhancer of \textbf{Sonic Hedgehog}.” In order to obtain a structured characterization of the sentence to be used in downstream applications, a NED system must map the  mention \textbf{Sonic Hedgehog} to the entity \textit{SHH gene}. To do so, the system can use context cues such as "enhancer" and "evolution", which commonly refer to genes, to avoid selecting semantically similar concepts such as \textit{Sonic Hedgehog protein} or \textit{Sonic Hedgehog signaling pathway}. 

Although NED systems have been successfully designed for general text corpora \cite{orr2020,yamada2020global,wu2020blink}, the NED task remains particularly challenging in the medical setting due to the presence of rare entities that occur infrequently in medical literature \cite{agrawal2020benchmarking}. As a knowledge-intensive task, NED requires the incorporation of structural resources, such as entity descriptions and category types, to effectively disambiguate rare entities \cite{orr2020}. However, this is difficult to accomplish in the medical setting for the following reasons: \smallskip
\begin{enumerate}
\item \textit{Coarse-grained and incomplete structural resources:} Metadata associated with entities in medical KBs is often coarse-grained or incomplete \cite{chen2009type,halper2011,agrawal2020benchmarking}. For example, over 65\% of entities in the United Medical Language System\footnote{\url{https://uts.nlm.nih.gov/uts/umls/home}} (UMLS) ontology, a popular medical KB, are associated with just ten types, suggesting that these types do not provide fine-grained disambiguation signals. In addition, over 93\% of entities in the UMLS KB have no associated description.

\item \textit{Low coverage over uncommon resources}:
Entities associated with some structural resources may occur infrequently in biomedical text. For instance, MedMentions~\cite{mohan2019medmentions}, which is one of the largest available biomedical NED datasets, contains fewer than thirty occurrences of entities with type “Drug Delivery Device”. In contrast, the high coverage type “Disease or Syndrome” is observed over 10,000 times. As a result, models may not learn effective reasoning patterns for disambiguating entities associated with uncommon structural resources, which limits the ability of the model to use these resources for resolving rare entities.\smallskip

\end{enumerate}
In this work, we design a biomedical NED system to improve disambiguation of rare entities through {\em cross-domain data integration}, which involves transferring knowledge between domains. Data integration across heterogeneous domains is a challenging problem with potential applications across numerous knowledge-intensive tasks. Here, we address this problem by utilizing a state-of-the-art general text entity linker to map medical entities to corresponding items in WikiData,\footnote{\url{https://www.wikidata.org/wiki/Wikidata:Main_Page}} a common general text KB. The key contributions of this work are listed below:\footnote{Code and data available at \url{https://github.com/HazyResearch/medical-ned-integration}.}\smallskip
\begin{itemize}
\item We generate structural resources for medical entities by incorporating knowledge from WikiData. This results in an augmented medical KB with a 12.8x increase in the number of entities with an associated description and a 2x increase in the average number of types for each entity. 
\item We utilize our integrated entity mappings to obtain pretraining datasets from PubMed and a medical subset of Wikipedia. These datasets include a total of 2.8M sentences annotated with over 4.2M entities across 23 thousand types. \smallskip
\end{itemize}

We evaluate our approach on two standard biomedical NED datasets: MedMentions and BC5CDR. Our results show that augmenting structural resources and pretraining across large datasets contribute to state-of-the-art model performance as well as up to a 57 point improvement in accuracy across rare entities that originally lack structural resources.

To the best of our knowledge, this is the first study to address medical NED through structured knowledge integration. Our cross-domain data integration approach can be translated beyond the medical domain to other knowledge-intensive tasks.

\section{Related Work}
\label{sec:related_work}
Recent state-of-the-art approaches for the medical NED task utilize transformer-based architectures to perform two tasks: candidate extraction, which involves identifying a small set of plausible entities, and reranking, which involves assigning likelihoods to each candidate. Prior methods for this task generally limit the use of structural resources from medical KBs due to missing or limited information \cite{bhowmik2021fast}. As a result, several existing approaches have been shown to generalize poorly to rare entities \cite{agrawal2020benchmarking}. Some previous studies have demonstrated that injecting auxiliary information, such as type or relation information, as well as pretraining can aid with model performance on various biomedical NLP tasks \cite{yuan2021kebiolm,liu2021sapbert,he2020infusing}. However, these works are limited by the insufficient resources in medical KBs as well as the use of pretraining datasets that obtain low coverage over the entities in the KB. Although some methods have been previously designed to enrich the metadata in medical ontologies with external knowledge, these approaches either use text-matching heuristics \cite{wang-etal-2018-ontology} or only contain mappings for a small subset of medical entities \cite{rahimi2020wikiumls}. Cross-domain structural knowledge integration has not been previously studied in the context of the medical NED task.  

\section{Methods}
\label{sec:methods}
We first present our cross-domain data integration approach for augmenting structural knowledge and obtaining pretraining datasets. We then describe the model architecture that we use to perform NED. 

\subsection{Cross-Domain Data Integration}
\label{sec:aug}
Rich structural resources are vital for rare entity disambiguation; however, metadata associated with entities in medical KBs is often too coarse-grained to effectively discriminate between textually-similar entities. We address this issue by integrating the UMLS Metathesaurus \cite{bodenreider2004}, which is the most comprehensive medical KB, with WikiData, a KB often used in the general text setting \cite{vrandecic2014wikidata}. We perform data integration by using a state-of-the-art NED system~\cite{orr2020} to map each UMLS entity to its most likely counterpart in WikiData; the canonical name for each UMLS entity is provided as input, and the system returns the most likely Wikipedia item. For example, the UMLS entity \textit{C0001621: Adrenal Gland Diseases} is mapped to the WikiData item \textit{Q4684717: Adrenal gland disorder}.

We then augment types and descriptions for each UMLS entity by incorporating information from the mapped WikiData item. For instance, the UMLS entity \textit{C0001621: Adrenal Gland Diseases} is originally assigned the type “Disease or Syndrome” in the UMLS KB; our augmentation procedure introduces the specific WikiData type “endocrine system disease". If the UMLS KB does not contain a description for a particular entity, we add a definition by extracting the first 150 words from its corresponding Wikipedia article. 

Our procedure results in an augmented UMLS KB with 24,141 types (190x increase). 2.04M entities have an associated description (12.8x increase). 

In order to evaluate the quality of our mapping approach, we utilize a segment of UMLS (approximately 9.3k entities) that has been previously annotated with corresponding WikiData items \cite{vrandecic2014wikidata}. Our mapping accuracy over this set is 80.2\%. We also evaluate integration performance on this segment as the proportion of predicted entities that share a WikiData type with the true entity, suggesting the predicted mapping adds relevant structural resources. Integration performance is 85.4\%. The remainder of items in UMLS have no true mappings to WikiData, underscoring the complexity of this task.

\subsection{Construction of Pretraining Datasets}
\label{sec:pretrain}
Existing datasets for the biomedical NED task generally obtain low coverage over the entities and structural resources in the UMLS knowledge base, often including less than 1\% of UMLS entities \cite{mohan2019medmentions}. Without adequate examples of structured metadata, models may not learn the complex reasoning patterns that are necessary for disambiguating rare entities. We address this issue by collecting the following two large pretraining datasets with entity annotations. Dataset statistics are summarized in Table \ref{tab:pretrain}.\smallskip

\begin{table}
\centering
\small
\begin{tabular}{lcc}
\hline
\textbf{} & \textbf{PubMedDS} & \textbf{MedWiki}\\
\hline
Total Documents & 508,295 & 813,541 \\
Total Sentences & 916,945 & 1,892,779\\
Total Mentions & 1,390,758 & 2,897,621\\ 
Unique Entities & 40,848 & 230,871\\
\end{tabular}
\caption{Dataset statistics for MedWiki and PubMedDS.}
\label{tab:pretrain}
\end{table}

\noindent
\textbf{MedWiki:} Wikipedia, which is often utilized as a rich knowledge source in general text settings, contains references to medical terms and consequently holds potential for improving performance on the medical NED task. We first annotate all Wikipedia articles with textual mentions and corresponding WikiData entities by obtaining gold entity labels from internal page links as well as generating weak labels based on pronouns and alternative entity names \cite{orr2020}. Then, we extract sentences with relevant medical information by determining if each WikiData item can be mapped to a UMLS entity using the data integration scheme described in \autoref{sec:aug}. 

MedWiki can be compared to a prior Wikipedia-based medical dataset generated by \citet{vashishth2021}, which utilizes various knowledge sources to map WikiData items to UMLS entities based on Wikipedia hyperlinks. When evaluated with respect to the prior dataset, our MedWiki dataset achieves greater coverage over UMLS, with 230k unique concepts (4x prior) and a median of 214 concepts per type (15x prior). However, the use of weak labeling techniques in MedWiki may introduce some noise into the entity mapping process (Section \ref{sec:aug} describes our evaluation of our mapping approach).\smallskip

\noindent
\textbf{PubMedDS:} The PubMedDS dataset, which was generated by \citet{vashishth2021}, includes data from PubMed abstracts. We remove all documents that are duplicated in our evaluation datasets.\smallskip 

We utilize the procedure detailed in Section \ref{sec:aug} to annotate all entities with structural information obtained from UMLS and WikiData. Final dataset statistics are included in Table \ref{tab:pretrain}. In combination, the two pretraining datasets include 2.8M sentences annotated with 267,135 unique entities across 23,746 types.   

\subsection{Model Architecture}
\label{sec:architecture}
We use a three-part approach for NED: candidate extraction, reranking, and post-processing.\par 
\noindent
\textbf{Candidate Extraction:} Similar to \citep{bhowmik2021fast}, we use the bi-encoder architecture detailed in \citet{wu2020blink} for extracting the top 10 candidate entities potentially associated with a mention. The model includes a context encoder, which is used to learn representations of mentions in text, as well as an entity encoder to encode the entity candidate with its associated metadata. Both encoders  are initialized with weights from SapBERT~\cite{liu2021sapbert}, a BERT model initialized from PubMedBERT and fine-tuned on UMLS synonyms. Candidate entities are selected based on the maximum inner product between the context and entity representations. We pretrain the candidate extraction model on MedWiki and PubMedDS. \smallskip

\noindent
\textbf{Reranking Model}: Given a sentence, a mention, and a set of entity candidates, our reranker model assigns ranks to each candidate and then selects the single most plausible entity. Similar to \citet{angell2020clustering}, we use a cross-encoder to perform this task. The cross-encoder takes the form of a BERT encoder with weights initialized from the context encoder in the candidate extraction model. \smallskip

\noindent
\textbf{Post-Processing (Backoff and Document Synthesis)}: Motivated by \citet{rajani2020explaining}, we back-off from the model prediction when the score assigned by the re-ranking model is below a threshold value and instead map the mention to the textually closest candidate. Then, we synthesize predictions for repeating mentions in each document by mapping all occurrences of a particular mention to the most frequently predicted entity. 

Further details about the model architecture and training process can be found in Appendix \ref{sec:app:model}.

\vspace{-0.1cm}
\section{Evaluation}
\label{sec:experiments}
We evaluate our model on two biomedical NED datasets and show that (1) our data integration approach results in state-of-the-art performance, (2) structural resource augmentation and pretraining are required in conjunction to realize improvements in overall accuracy, and (3) our approach contributes to a large performance lift on rare entities with limited structural resources.

\subsection{Datasets}
\label{sec:eval:datasets}
We evaluate our model on two NED datasets, which are detailed below. Additional dataset and preprocessing details can be found in Appendix \ref{sec:app:data}.

\noindent
\begin{itemize}
\item \textbf{MedMentions (MM)} is one of the largest existing medical NED datasets and contains 4392 PubMed abstracts annotated with 203,282 mentions. We utilize the ST21PV subset of MM, which comprises a subset of concepts deemed by the authors to be most useful for semantic indexing.
\item \textbf{BC5CDR} contains 1500 PubMed abstracts annotated with 28,785 mentions of chemicals and diseases \cite{li2016}.
\end{itemize}
\noindent

We use all available UMLS structural resources when preprocessing datasets, and as a result, we map MM entities to 95 UMLS types and BC5CDR entities to 47 UMLS chemical and disease types.

\begin{table}
\small
\centering
\resizebox{0.47\textwidth}{!}{%
\begin{tabular}{lcc}
\hline
\textbf{} & \textbf{MM} & \textbf{BC5CDR}\\
\hline
\citet{bhowmik2021fast} &   68.4
 & 84.8 \\
\citet{angell2020clustering} & 72.8 & 90.5 \\
Ours (Full) & \textbf{74.6$_{\pm0.1}$} & \textbf{91.5$_{\pm0.1}$} \\
\midrule
\citet{angell2020clustering}+Post-Processing & 74.1 & 91.3\\
Ours+Post-Processing & \textbf{74.8$_{\pm0.1}$} & \textbf{91.9$_{\pm0.2}$} \\
\end{tabular}}
\caption{\textit{Benchmark Performance.} We compare performance of our model to prior work. Metrics indicate accuracy on the test set. We report the mean and standard deviation across five training runs.\smallskip}
\label{tab:results}
\end{table}

\begin{table}
\small
\centering
\begin{tabular}{lcc}
\hline
\textbf{} & \textbf{MM} & \textbf{BC5CDR}\\
\hline
Ours (Baseline) & 74.0$_{\pm0.2}$ & 89.3$_{\pm0.1}$\\
Ours (Augmentation Only) & 74.1$_{\pm0.1}$ & 89.3$_{\pm0.1}$ \\
Ours (Full) & 74.6$_{\pm0.1}$ & 91.5$_{\pm0.1}$\\
\end{tabular}
\caption{\textit{Model Ablations.} We measure accuracy of our full model (Full), our model with augmented structural resources and no pretraining (Augmentation Only), and our model without augmented structural resources and without pretraining (Baseline). We report the mean and standard deviation across five training runs.}
\label{tab:ablations}
\end{table}

\subsection{Performance on Benchmarks}
We compare our approach to prior state-of-the-art methods from \citet{bhowmik2021fast}\footnote{\citet{bhowmik2021fast} uses the complete MM dataset, while \citet{angell2020clustering} and our work use the MM-ST21PV subset.} and \citet{angell2020clustering}. As shown in \autoref{tab:results}, our approach with post-processing\footnote{Note that our post-processing method (\autoref{sec:architecture}) differs from the post-processing method used in \citet{angell2020clustering}.} sets a new state-of-the-art on MM by 0.7 accuracy points and BC5DR by 0.6 points. In addition, our method without post-processing (Full) outperforms comparable methods by up to 1.8 accuracy points.

\subsection{Ablations}
In order to measure the effect of our data integration approach on model performance, we perform various ablations as shown in \autoref{tab:ablations}. We find marginal performance improvement when augmented structural resources are used without pretraining (Augmentation Only Model). When pretraining and augmented structural resources are used in conjunction (Full Model), we observe a performance lift on both datasets, suggesting that the model can only learn fine-grained reasoning patterns when both components are incorporated into the model. 

We observe that our approach leads to a larger improvement on BC5CDR (2.2 points) than MM (0.6 points). The lack of overall improvement for the MM dataset is expected, since the original MM dataset consists of finer-grained types than the BC5CDR dataset. Specifically, we observe that 95\% of the entities in BC5CDR are categorized with just 15 types, and in comparison, only 57\% of entities in MM can be categorized with 15 types. This suggests that the magnitude of model improvement is
likely to be dependent on the original granularity of structural resources in the training dataset. As a result, our data integration approach will naturally yield greater performance improvements on the BC5CDR dataset.

\begin{figure}
    \centering
    \includegraphics[width=0.5\textwidth] {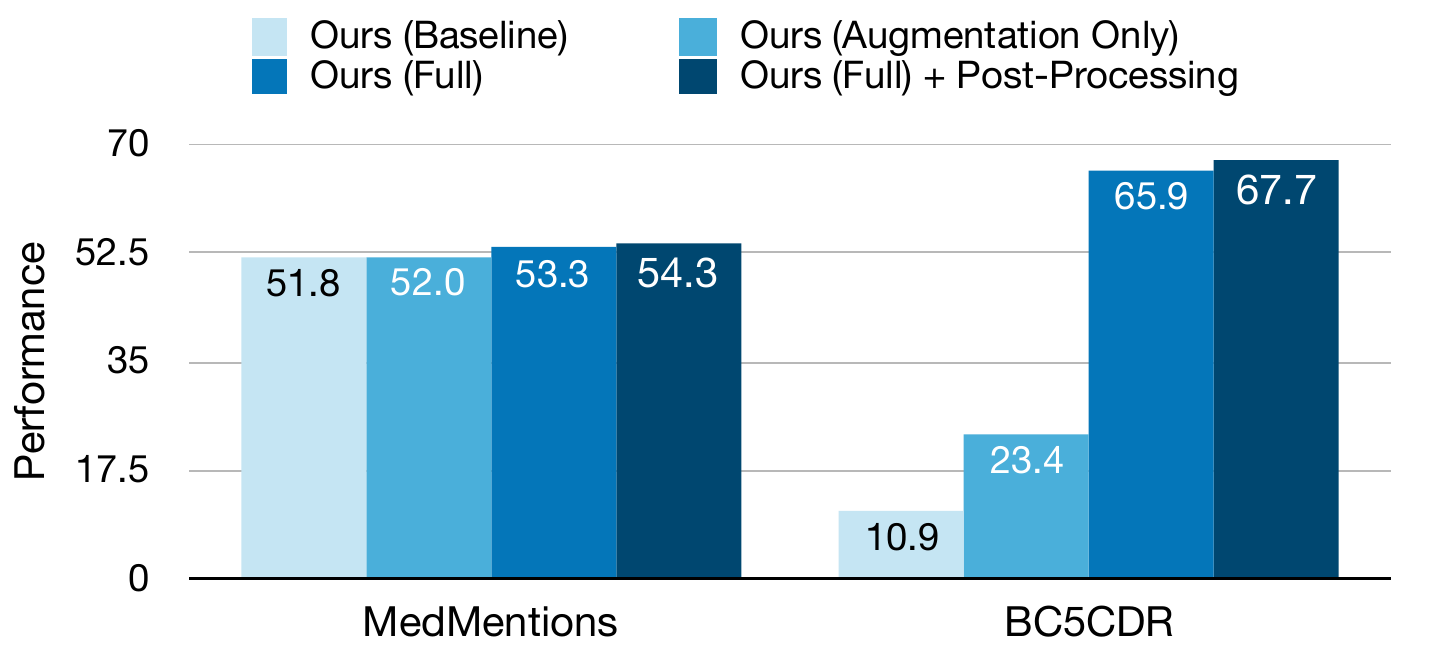}
    \caption{\textit{Performance on Rare Entities with Limited Structural Resources.} We measure the test accuracy of four ablation models on a subset of rare entities that have limited structural resources. We report mean values across five training runs.}
    \label{fig:rare}
\end{figure}

\subsection{Performance on Rare Entities}
In \autoref{fig:rare}, we measure performance on entities that appear less than five times in the training set and are associated with exactly one type and no definition in the UMLS KB. We observe an improvement of 2.5 accuracy points on the MM dataset and 56.8 points on BC5CDR. Results on the BC5CDR dataset also show that utilizing pretraining and resource augmentation in combination leads to a 3x improvement in performance when compared to the Augmentation Only model; this further supports the need for both pretraining and structural resource augmentation when training the model. We observe similar trends across entities with limited metadata that never appear in pretraining datasets. Additional evaluation details are included in Appendix \ref{sec:app:eval}.

\section{Conclusion}
In this work, we show that cross-domain data integration helps achieve state-of-the-art performance on the named entity disambiguation task in medical text. The methods presented in this work help address limitations of medical knowledge bases and can be adapted for other knowledge-intensive problems.

\section*{Acknowledgements}
We are thankful to Sarah Hooper, Michael Zhang, Dan Fu, Karan Goel, and Neel Guha for helpful discussions and feedback. MV is supported by graduate fellowship awards from the the Department of Defense (NDSEG) and the Knight-Hennessy Scholars program at Stanford University. LO is supported by an IC Postdoctoral Fellowship. ML is supported by an NSF Graduate Research Fellowship (No. DGE-1656518). We gratefully acknowledge the support of NIH under No. U54EB020405 (Mobilize), NSF under Nos. CCF1763315 (Beyond Sparsity), CCF1563078 (Volume to Velocity), and 1937301 (RTML); ONR under No. N000141712266 (Unifying Weak Supervision); ONR N00014-20-1-2480: Understanding and Applying Non-Euclidean Geometry in Machine Learning; N000142012275 (NEPTUNE); the Moore Foundation, NXP, Xilinx, LETI-CEA, Intel, IBM, Microsoft, NEC, Toshiba, TSMC, ARM, Hitachi, BASF, Accenture, Ericsson, Qualcomm, Analog Devices, the Okawa Foundation, American Family Insurance, Google Cloud, Salesforce, Total, the HAI Cloud Credits for Research program, the Stanford Data Science Initiative (SDSI), and members of the Stanford DAWN project: Facebook, Google, and VMWare. The Mobilize Center is a Biomedical Technology Resource Center, funded by the NIH National Institute of Biomedical Imaging and Bioengineering through Grant P41EB027060. The U.S. Government is authorized to reproduce and distribute reprints for Governmental purposes notwithstanding any copyright notation thereon. Any opinions, findings, and conclusions or recommendations expressed in this material are those of the authors and do not necessarily reflect the views, policies, or endorsements, either expressed or implied, of NIH, ONR, or the U.S. Government.

\bibliography{anthology}
\bibliographystyle{acl_natbib}

\clearpage
\appendix
\section{Appendix}
\label{sec:appendix}
\subsection{Data Details}
\label{sec:app:data}
\subsubsection{UMLS Knowledge Base}
We utilize the 2017 AA release of the UMLS Metathesaurus as the KB, filtered to include entities from 18 preferred source vocabularies \cite{mohan2019medmentions,bodenreider2004}. The dataset includes 2.5M entities associated with 127 types. Approximately 160K entities have an associated description. 

\subsubsection{Construction of Pretraining Datasets}
We obtain two pretraining datasets: MedWiki and PubMedDS. After collecting each dataset using the methods detailed in Section \ref{sec:pretrain}, we downsampled to address class imbalance between entities, since some entities were represented at higher rates than others. Sentences were removed if all entities within the sentence were observed in the dataset with high frequency (defined as occurring in at least 40 other sentences).

Prior work by \cite{newman2020ambiguity} demonstrates the importance of including ambiguity in in medical NED training datasets. \citet{newman2020ambiguity} defines dataset ambiguity as the number of unique entities associated with a particular mention string. By this definition, the MedWiki training set has 25k ambiguous mentions (7\% of unique mentions), with a minimum, median, and maximum ambiguity per mention of 2.0, 2.0, and 29.0 respectively. PubMedDS includes 7.6k ambiguous mentions (36\% of unique mentions), with a minimum, median, and maximum ambiguity per mention of 2.0, 3.0, and 24.0 respectively.

\subsubsection{Evaluation Datasets}
We evaluate our model across two medical NED benchmark datasets: MedMentions and BC5CDR. Dataset and preprocessing details are provided below.

\noindent
\textbf{MedMentions (MM)} \cite{mohan2019medmentions}: MM consists of text collected from 4392 PubMed abstracts. We use all available UMLS structural resources when preprocessing datasets, and as a result, we map MM entities to 95 UMLS types.

We preprocess the dataset by (1) expanding abbreviations using the Schwartz-Hearst algorithm \cite{schwartz_hearst_2003}, (2) splitting documents into individual sentences with the Spacy library, (3) converting character-based mention spans to word-based mention spans, and (4) grouping sentences into sets of three in order to provide adequate context to models. Mentions occurring at sentence boundaries, overlapping mentions, and mentions with invalid spans (when assigned by the Spacy library) are removed from the dataset during pretraining, resulting in a total of 121K valid mentions in the training set, 8.6K mentions in the validation set, and 8.4K mentions in the test set. Preprocessed dataset statistics are summarized in Table \ref{tab:medmentions}. 

\noindent
\textbf{BC5CDR} \cite{li2016}: BC5CDR consists mentions mapped to chemical and disease entities. Entities are labeled with MESH descriptors; MESH is a medical vocabulary that comprises a subset of the UMLS KB. 

We preprocess the dataset by (1) expanding abbreviations using the Schwartz-Hearst algorithm \cite{schwartz_hearst_2003}, (2) splitting all composite mentions into multiple parts, (3) splitting documents into individual sentences with the Spacy library, (4) converting character-based mention spans to word-based mention spans, and (5) grouping sentences into sets of three in order to provide adequate context to models. Composite mentions that could not be separated into multiple segments were removed from the dataset; mentions with MESH descriptors that were missing from the 2017 release of the UMLS KB were also removed. This resulted in a total of 9257 valid mentions in the training set, 1243 mentions in the validation set, and 1300 mentions in the test set. Preprocessed dataset statistics are summarized in \autoref{tab:BC5CDR}.


\begin{table}
\centering
\small
\begin{tabular}{lccc}
\hline
\textbf{} & \textbf{Train} & \textbf{Dev} & \textbf{Test}\\
\hline
Total Documents & 2635 & 878 & 879\\
Total Sentences & 9008 & 2976 & 2974\\ 
Total Mentions & 121,861 & 40,754 & 40,031\\ 
Unique Entities & 18,495 & 8637 & 8449\\
\end{tabular}
\caption{Dataset statistics for MedMentions after preprocessing.}
\label{tab:medmentions}
\end{table}

\begin{table}
\centering
\small
\begin{tabular}{lccc}
\hline
\textbf{} & \textbf{Train} & \textbf{Dev} & \textbf{Test}\\
\hline
Total Documents & 500 & 500 & 500\\
Total Sentences & 1431 & 1431 & 1486\\
Total Mentions & 9257 & 9452 & 9628\\ 
Unique Entities & 1307 & 1243 & 1300\\
\end{tabular}
\caption{Dataset statistics for BC5CDR after preprocessing.}
\label{tab:BC5CDR}
\end{table}

\subsection{Model Details}
\label{sec:app:model}
We now provide details of our bi-encoder candidate generator, cross-encoder re-ranker, and post-processing method.

\begin{table}[]
    \centering
    \begin{tabularx}{3.0in}{|X|c|c|}
        \hline
        Param & Bi-encoder & Cross-encoder \\ \hline
        learning rate & $1e^{-5}$ & $2e^{-5}$ \\ \hline
        weight decay & $0$ & $0.01$ \\ \hline
        $\beta_1$ & $0.9$ & $0.9$ \\ \hline
        $\beta_2$ & $0.999$ & $0.999$ \\ \hline
        $\mathit{eps}$ & $1e^{-6}$ & $1e^{-6}$ \\ \hline
        effective batch size & $100$ & $128$ \\ \hline
        epochs & 3-10 & 10 \\ \hline
        warmup & 10\% & 10\% \\ \hline
        learning rate scheduler & linear & linear \\ \hline
        optimizer & AdamW & AdamW \\ 
        \hline
    \end{tabularx}
    \caption{Learning Parameters for the bi-encoder and cross-encoder}
    \label{tab:learning_params}
\end{table}

\subsubsection{Candidate Generation with a Bi-encoder}
Given a sentence and mention, our candidate generator model selects which top $K$ candidates are the most likely to be the entity referred to by the mention. Similar to \citep{bhowmik2021fast}, we use a BERT bi-encoder to jointly learn representations of mentions and entities. The bi-encoder has a context encoder to encode the mention and an entity encoder to encode the entity. The candidates are selected based on those that have the highest maximum inner product with the mention representation.

The context tokenization is
$$
\texttt{[CLS]} c_{\ell} \texttt{[ENT\_START]} m \texttt{[ENT\_END]} c_{r} \texttt{[SEP]}
$$
where $\texttt{[ENT\_START]}$ and $\texttt{[ENT\_END]}$ are new tokens to indicate where the mention is in the text. We set the left and right window length to be 30 words with the max tokens used for the sentence tokens of 64.

The entity tokenization is
$$
\texttt{[CLS]} \mathit{title} \texttt{[SEP]} \mathit{types} \texttt{[SEP]} \mathit{desc} \texttt{[SEP]}
$$
where $\mathit{title}$ is the entity title, $\mathit{types}$ is a semi-colon separated list of types, and $\mathit{desc}$ is the description of an entity. We limit the list of types such that the total length of $\mathit{types}$ is less than 30 words. The max length for the entity tokens is 128. This means that the description may be truncated if it exceeds the maximum length.

\paragraph{Training}
We train the bi-encoder similar to \cite{wu2020blink}. We run in three phases. The first is where all negatives are in-batch negatives with a batch size of 100. The next two phases take the top 10 predicted entities for each training example as additional negatives for the batch with a batch size of 10. Before each phase, we re-compute the 10 negatives.

For pretraining, we run each phase for 3 epochs. When fine-tuning on specific datasets, we run each for 10 epochs. All training parameters are shown in \autoref{tab:learning_params}.

During pretraining, candidates are drawn from the entire UMLS KB, consisting of 2.5M entities. During fine-tuning on the MM dataset, candidates are drawn from the valid subset of entities defined in the ST21PV version of the dataset, which includes approximately 2.36M entities. During fine-tuning on the BC5CDR dataset, candidates are drawn from a set of 268K entities with MESH identifiers.

\subsubsection{Reranker Cross-encoder}
Given a sentence, mention, and a set of entity candidates, our reranker model selects which candidate is the most likely entity referred to by the mention. Similar to \citet{angell2020clustering}, we use a BERT cross-encoder architecture to learn a score for each entity candidate --- mention pair. The models takes as input the sequence of tokens
$$
\mathit{context} \texttt{[ENT\_DESC]} \mathit{entity}
$$
where $\mathit{context}$ is the context tokenization from the bi-encoder, $\mathit{entity}$ is the entity tokenization from the bi-encoder, and $\texttt{[ENT\_DESC]}$ is a special tag to indicate when the entity description is starting. One difference from the bi-encoder is that the title of the entity includes the canonical name as well as all alternate names. We keep the length parameters the same as for the bi-encoder except we let the context have a max length of 128. We take the output representation from the \texttt{[CLS]} token and project it to a single dimension output. We pass the outputs for each candidate through a softmax to get a final probability of which candidate is most likely. 

\paragraph{Training}
When training the cross encoder, we warm start the model with the context model weights from the candidate generator bi-encoder. We train all models using the top 10 candidates, and we train for 10 epochs. 
We use standard fine-tuning BERT parameters, shown in \autoref{tab:learning_params}. 

We do not separately pretrain the cross encoder on our pretraining datasets. Pretrained knowledge is instead transferred through the use of context encoder weights for warm starting the model.

\subsubsection{Post-Processing (Backoff and Document Synthesis)}
We post-process model outputs by backing off from the model prediction when the score assigned by the re-ranking model is below a threshold value. We utilize the validation set to determine the optimal value of the threshold, which we select as 0.55 for MM and 0.45 for BC5CDR. 

Then, we group predictions for each document, which ensures that all repeating mentions in a document will map to the same entity. We map each occurrence of a repeating mention within a document to the most frequently-predicted entity. For example, assume that the mention "DFS" occurs three times in a document, with the occurrences resolved to the entities "Diabetic Foot Ulcer", "Diabetic Foot Ulcer", and "DF 118". In this case, we assign the most frequent prediction, which is "Diabetic Foot Ulcer", to all occurrences of the mention DFU.

\subsection{Extended Evaluations}
\subsubsection{Candidate Generation Performance}

\begin{table}
\small
\centering
\begin{tabular}{lcc}
\hline
\textbf{} & \textbf{MM} & \textbf{BC5CDR}\\
\hline
\cite{bhowmik2021fast} & -- / 87.6 & -- / 92.3\\
\cite{angell2020clustering} & 50.8 / <82.3 &  86.9 / <93.1\\
\midrule
Ours (Baseline) & 70.1 / 88.4 & 83.5 / 93.3\\ 
Ours (Augmentation Only) & 70.3 / \textbf{88.5} & 83.7 / 93.1 \\ 
Ours (Full) & \textbf{71.7} / 88.3 & \textbf{89.2} / \textbf{96.2}\\
\end{tabular}
\caption{\textit{Performance of Candidate Generator on MM and BC5CDR (Recall@1 / Recall@10).} Our approach leads to improvements in candidate recall.} 
\label{tab:candgen}
\end{table}

\autoref{tab:candgen} shows performance of our candidate generation approach and compares against \citep{angell2020clustering} and \citep{bhowmik2021fast}. Note that \citep{bhowmik2021fast} also uses a bi-encoder for candidate generation. As in \autoref{tab:ablations}, we ablate the three models without augmentation or pretraining (Baseline), with augmentation only (Augmentation Only), and with augmentation and pretraining (Full).

We find our method outperforms both prior works in Recall@1 and Recall@10. We further find similar trends as in \autoref{tab:ablations} where augmentation without pretraining provides a limited lift of 0.2 accuracy points in Recall@1 performance. With pretraining, we see a more substantial lift of 1.6 points on MM and a 5.7 points on BC5CDR.

\subsubsection{Evaluation on Subpopulations}
\label{sec:app:eval}

\newcolumntype{m}{>{\hsize=.5\hsize}X}
\begin{table}[]
\begin{small}
    \centering
    \begin{tabularx}{3.3in}{|m|X|}
        \cline{1-2}
        Subpopulation & Description \\ \cline{1-2}
        Multi- (Single) Word Mentions & Mentions that are multiple (single) words \\ \cline{1-2}
        Unseen Mentions & Mentions that are unseen in fine-tuning training \\ \cline{1-2}
        Unseen Entities & Entities that are unseen in fine-tuning training \\ \cline{1-2}
        Not Direct Match & Mentions that are not a preferred name or synonym of the entity \\ \cline{1-2}
        Top 100 & Mentions that are mapped to the top 100 entities in fine-tuning data \\ \cline{1-2}
        Unpopular & Mentions that are more commonly mapped to a different entity \\ \cline{1-2}
        Limited Metadata & Entities that have no description and only one UMLS type before augmentation \\ \cline{1-2}
        Rare \& Limited Metadata & Limited metadata entities that appear less than 5 times in fine-tuning data\\ \cline{1-2}
        Never Seen \& Limited Metadata & Limited metadata entities that do not appear in pretraining data or fine-tuning data\\ \cline{1-2}
    \end{tabularx}
    \caption{Subpopulations used to compare models. Each model's accuracy is measured on the subset of data defined for each subpopulation.}
    \label{tab:subpops}
\end{small}
\end{table}

\begin{figure*}[ht]
\centering
\includegraphics[width=\linewidth]{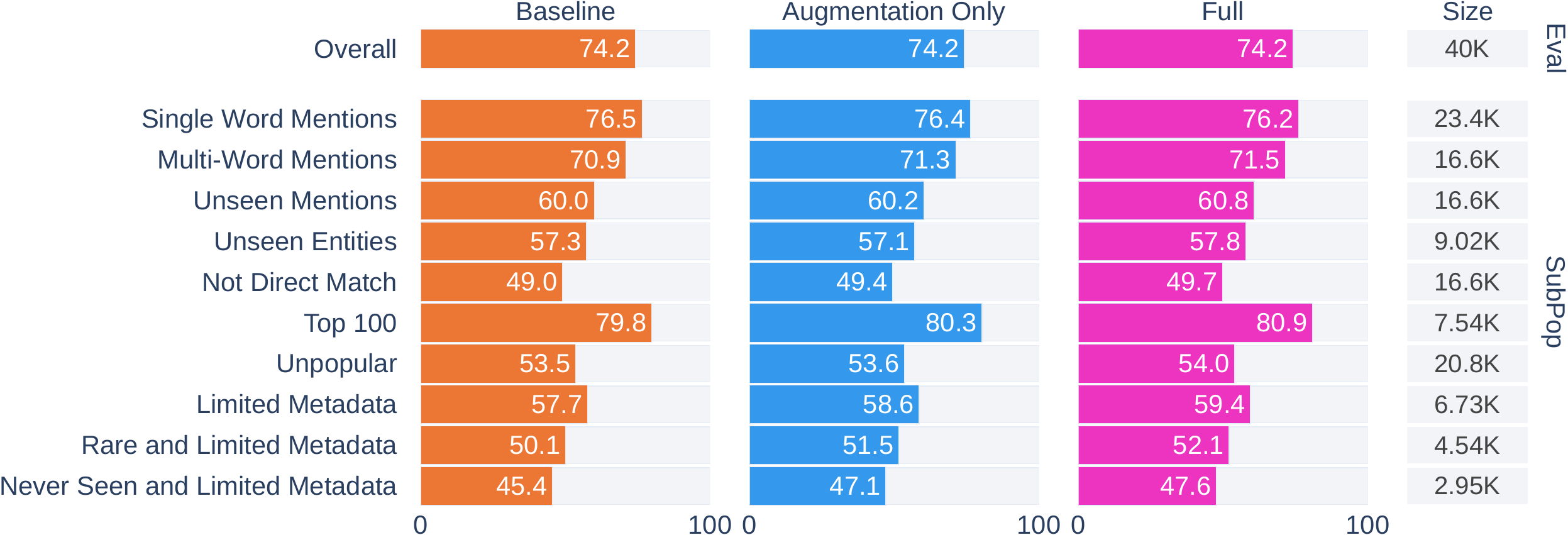}
\caption{\label{fig:mm_rg}
Accuracy over subpopulations for our three ablation models on MedMentions. 
}
\end{figure*}

\begin{figure*}[ht]
\centering
\includegraphics[width=\linewidth]{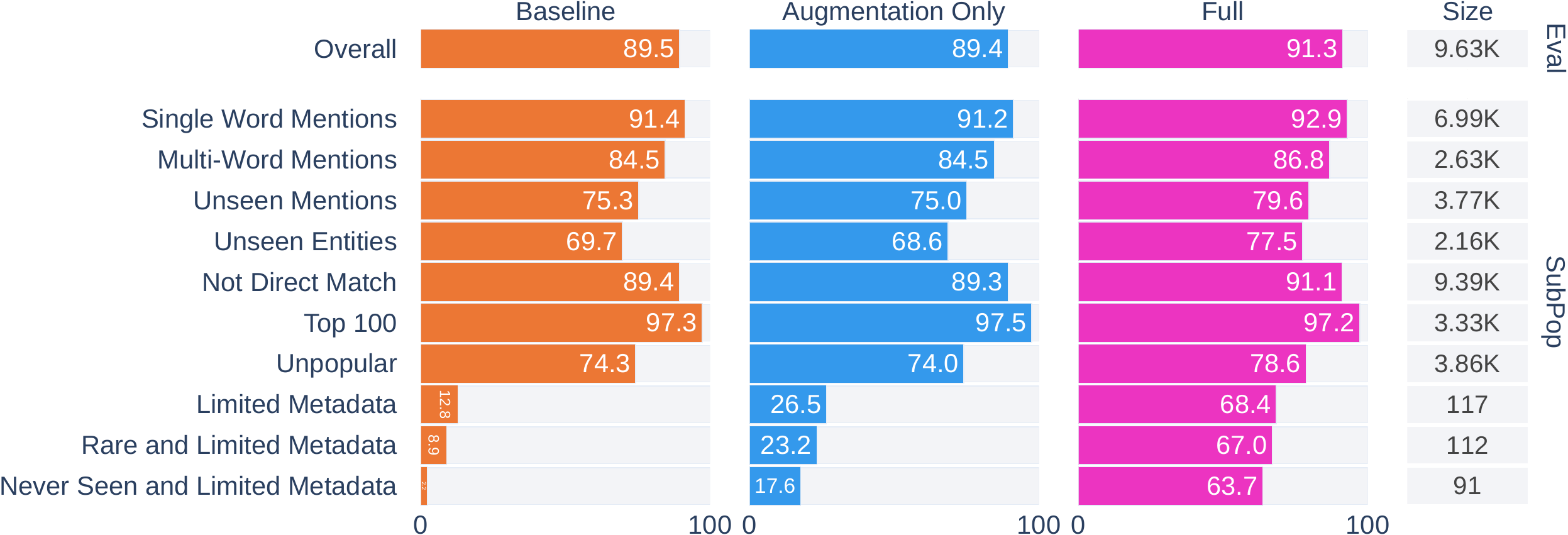}
\caption{\label{fig:bc5_rg}
Accuracy over subpopulations for our three ablation models on BC5CDR. 
}
\end{figure*}

For fine-grained analysis of all models, we use the Robustness Gym toolkit~\cite{goel2021robustness} to create relevant subpopulations to measure model accuracy. \autoref{tab:subpops} describes the subpopulations we use for evaluation. We take those described in \citep{agrawal2020benchmarking} as well as custom ones we used in \autoref{sec:experiments}.

\autoref{fig:mm_rg} and \autoref{fig:bc5_rg} show the performance on MedMentions and BC5CDR across the subpopulations. We note the following trends.

\begin{itemize}
    \item \textit{Never seen entities rely on pretrained structural resources}. When looking at the subpopulation of entities that are not seen in pretraining data or fine-tuning data, we see a 1.7 accuracy point lift in MM and 15 point lift in BC5CDR just from adding augmented resources. This is further improved by 0.5 points in MM and 46 points in BC5CDR. As these entities are never seen during training, the improvement from pretraining likely comes from the improved ability of the model to reason over the structural resources. 
    
    \item \textit{Popular entities achieve the highest performance}. Unsurprisingly, across both datasets, we see the largest evaluation accuracy scores (up to 80.9 and 97.2 for MM and BC5CDR respectively) on subpopulations where the entity is one of the 100 most popular in the training dataset. Since these entities occur repeatedly during training, the model is able to memorize relevant disambiguation patterns.
    
    \item \textit{Unseen entities are easier to resolve than rare entities with limited structural metadata}. Unseen entities are those that are not seen by the model during training. As a result, these are typically considered the most difficult entities to resolve~\cite{orr2020, logeswaran-2019-zero}. We find that across both datasets and all models, the ``Rare and Limited Metadata`` subpopulation performs up to 61 accuracy points worse than the unseen entity slicing. This further supports the need for structural metadata when resolving rare entities.
    
    \item \textit{There is a significant performance gap between two datasets on the the ``Not Direct Match`` slice}. We find that performance on the ``Not Direct Match`` MM subpopulation is up to 41 accuracy points lower than the same subpopulation  in BC5CDR. 
\end{itemize}

\end{document}